\documentclass[runningheads]{llncs}
\usepackage{graphicx,wrapfig}
\usepackage{amsmath}
\usepackage{algorithm}
\usepackage{subfigure}
\usepackage{multirow}
\usepackage{amssymb}
\usepackage{booktabs,array,dcolumn}
\usepackage[noend]{algpseudocode}

\begin{document}
\title{Hyp-RL : Hyperparameter Optimization by Reinforcement Learning\thanks{The authors would like to thank Nicolas Schilling for his valuable advice.}
}
%
%\titlerunning{Abbreviated paper title}
% If the paper title is too long for the running head, you can set
% an abbreviated paper title here
%
\author{Hadi S. Jomaa\inst{1}\and%\orcidID{} \and
Josif Grabocka\inst{1}\and %\orcidID{} \and
Lars Schmidt-Thieme\inst{1}}%\orcidID{}}
\authorrunning{H.S. Jomaa et al.}
% First names are abbreviated in the running head.
% If there are more than two authors, 'et al.' is used.
%
\institute{University of Hildesheim, Universit\"atsplatz 1, 31141 Hildesheim, Germany 
\email{\{jomaah,josif,lars\}@ismll.uni-hildesheim.de}}
\maketitle              % typeset the header of the contribution
\begin{abstract}
Hyperparameter tuning is an omnipresent problem in machine learning
as it is an integral aspect of obtaining the state-of-the-art performance
for any model. Most often, hyperparameters are optimized just by training 
a model on a grid of possible hyperparameter values and taking the one
that performs best on a validation sample (grid search). More recently,
methods have been introduced that build a so-called surrogate model that 
predicts the validation loss for a specific hyperparameter setting, model 
and dataset and then sequentially select the next hyperparameter to test, 
based on a heuristic function of the expected value and the uncertainty of 
the surrogate model called acquisition function (sequential model-based 
Bayesian optimization, SMBO).
\\
In this paper we model the hyperparameter optimization problem as a 
sequential decision problem, which hyperparameter to test next, 
and address it with reinforcement learning. This way our model does
not have to rely on a heuristic acquisition function like SMBO, but
can learn which hyperparameters to test next based on the subsequent
reduction in validation loss they will eventually lead to, either 
because they yield good models themselves or because they allow the
hyperparameter selection policy to build a better surrogate model
that is able to choose better hyperparameters later on. Experiments
on a large battery of 50 data sets demonstrate that our method 
outperforms the state-of-the-art approaches for hyperparameter learning.

\end{abstract}
\section{Introduction}
Hyperparameter tuning plays a significant role in the overall performance of machine learning models and can be the main factor in deciding whether a trained model turns out to be the state-of-the-art or simply moderate \cite{wistuba2016two}. %Unfortunately, there seems to be inconsistencies within the related literature when it comes to the optimal hyperparameter configurations as well as a range of missing values that often go unreported \cite{henderson2018deep}.
The growing sizes of data sets and the complexities of the models result in an increased training time and require more resources, which makes hyperparameter optimization an even harder task. When automating hyperparameter tuning, practical solutions must be robust, scalable and maintain low computational budget, while ensuring a strong final performance \cite{falkner2018bohb}.
Several solutions have been proposed to tune hyperparameters over the years which vary in terms of computational complexity and scalability. Some more traditional approaches include manual search, grid search, and random search \cite{bergstra2012random}, whereas others, such as Bayesian optimization techniques \cite{shahriari2016taking,hutter2011sequential}, rely on probabilistic models, most commonly Gaussian processes, to optimize the expected improvement of the surrogate model, by treating the response function (e.g. validation loss) as a black box function. %Meta-learning approaches \cite{van2015fast,gomes2012combining}. Nevertheless, these 'black boxes' \cite{chang2011libsvm} that are conventionally used, do not offer insight into the evolution of the training process, as they aim to map the hyperparameter space in a gradient-free approach.
\\
The search for the best hyperparameter configuration is a sequential decision process \cite{xu2017reinforcement} in which initial values are set, and later adjusted, through a mixture of intuition and trial-and-error, to optimize an observed performance on typically the hold-out validation set, i.e. to maximize the accuracy or minimize the loss. Hence, finding a scalable and efficient policy for hyperparameter tuning is critical, especially for high-dimensional function estimators, such as deep neural networks that require longer time to evaluate a single hyperparameter response. To achieve an automatic hyperparameter tuning, one must also take into account how \textit{well} certain configurations performed on other data sets and carry this knowledge over to new data sets. A well trained policy would then be able to navigate the response surface of a learning algorithm based on previous experiences, in order to converge rapidly to a global optimum. 
%It is therefore possible to use a controller that automatically navigates the hyperparameter space based on the observed hyperparameter settings and data set-specific meta features. 
\\
In this work, we pose hyperparameter tuning within a reinforcement learning (RL) framework, where the agent learns to explore the hyperparameter space of a fixed network topology. The task in RL is to optimize an agent's return over a fixed budget of episodes, by learning to estimate the expected return at a given state, and optimizing the action selection process in a way that maximizes the return \cite{xu2018meta}. RL research has witnessed rapid progress within the last decade \cite{henderson2018deep}, showing promising results in several areas such as robotic control \cite{lillicrap2015continuous,schulman2017proximal}, game playing \cite{mnih2015human,silver2016mastering,silva2017dynamic}, e-commerce \cite{cai2018reinforcement,zhao2018impression}, and more. The core of hyperparameter optimization boils down to selecting the optimal configuration, given an observable history of configurations and corresponding rewards, in a way that increases future rewards, which falls seemlessly into a RL framework. In this paper, we learn a controller that can tune the hyperparameters of a fixed topology by defining a state representation, set of actions, and a transition function which allow an agent to navigate the response surface and maximize its reward.
\\
We summarize the contributions of the paper as follows:
\begin{itemize}
\item This is the first paper, to the best of our knowledge, that formulizes hyperparameter tuning as a reinforcement learning problem, and provides empirical evidence on its feasibility
\item A novel policy based on Q-learning, Hyp-RL, that can navigate high-dimensional hyperparameter spaces
\item Formulation of hyperparameter response surface as an environment with a defined reward function and action space
\item Experiments on a meta-data set created using 50 randomly selected UCI classification data sets that highlight the strong final performance of the controller and scalability across varying hyperparameter response surfaces
\item Empirical evaluation that showcases the consistent gain in classification performance based on the initial hyperparameter trials selected by our controller
\end{itemize} 
%We summarize the contributions of the paper as follows:
%\begin{itemize}
%\item This is the first paper, to the best of our awareness, that formulizes hyperparameter tuning as a reinforcement learning problem, and provides empirical evidence on its feasibility%A novel hyperparameter controller based on Q-learning that can navigate high-dimensional hyperparameter response surfaces across data sets, making use of knowledge transfer
%\item Experiments on a meta-data set created using 50 randomly selected UCI classification data sets that highlight the strong final performance of the controller and scalability across varying hyperparameter response surfaces
%\item Empirical evaluation that showcases the consistent gain in classification performance of the based on the initial hyperparameter trials selected by our controller
%\end{itemize} 
\section{Related Work}
Hyperparameter optimization is still considered an open problem within the machine learning community, despite being widely used in practice to improve the performance of any learning algorithm. Perhaps the simplest approaches include manual search, selecting configurations based on intuition, grid search, which allocates equal weights to unimportant parameters, and random search \cite{bergstra2012random} that suffers from the lack of guidance.
\\
Bayesian optimization techniques \cite{shahriari2016taking} model the hyperparameter space in a probabilistic framework and obtain smooth uncertainty estimates when modelling the validation error. Despite the demonstrated powerful performance on tuning convolutional \cite{snoek2012practical,komer2014hyperopt} and fully-connected \cite{mendoza2016towards} neural networks, these methods are based on Gaussian processes that require an increasing number of data points with higher dimensional spaces and setting hyperpriors \cite{falkner2018bohb}. 
Other approaches have also been proposed as a combination of existing methods, building on top of their existing merits \cite{falkner2018bohb,snoek2015scalable}.
\\
Meta-knowledge is also leveraged to accelerate hyperparameter tuning across data sets. The straightforward way is through a warm-start initialization which can extend to all methods. This has led to a significant improvement for Bayesian optimization techniques based on Gaussian processes \cite{wistuba2015sequential,feurer2015initializing,lindauer2018warmstarting} and on neural networks \cite{perrone2018scalable}. The transfer of surrogate models, limited to the sequential model-based optimization \cite{jones1998efficient} framework is also explored, where independent surrogate models are either combined in an ensemble \cite{wistuba2018scalable,wistuba2016two,feurer2018scalable}, or a surrogate model is trained across data sets by modeling a common response surface \cite{yogatama2014efficient}.
\\
Hyperparameter optimization is also addressed within the scope of reinforcement learning, specifically for architectural network design. In \cite{zoph2016neural}, an RNN-based policy network is proposed which iteratively generates new architectures based on the gradient information from its child networks. At the end of the training process, the model that resulted in the best validation error is selected for further training. A meta-modeling approach is also proposed within a Q-learning setting \cite{baker2016designing}. The final architecture produces by these agents achieved state-of-the-art performance on several image classification data sets. For online learning rate adjustments, gradient information are used as state representation at every step \cite{xu2017reinforcement}. Perhaps the most similar work to our proposed approach is hyperparameter optimization for tracking applications ~\cite{DBLP:conf/cvpr/DongSWL0P18}. Using a continuous deep Q-learning network and a sequence of images as a state representation, hyperparameters are updated to maximize the performance of the tracker.
\\
In this paper, we cast the problem of hyperparameter tuning into a reinforcement learning task. We define an environment where an agent learns to navigate the hyperparameter space and arrives at the best configuration, through a series of actions representing the hyperparameter configuration to be evaluated, and observes the reward provided by the response function. Hyperparameter configurations are sequentially chosen by a policy function that replaces the acquisition function utilized in Bayesian optimization based approaches and balances between exploration and exploitation. Posing the problem in this fashion allows to exploit all the bells and whistles of reinforcement learning, and to the best of our knowledge, we are the first to formulate hyperparameter tuning in this manner.
\section{Problem Definition: Hyperparameter Optimization}
The objective of a machine learning algorithm $\mathcal{A} : \mathcal{D}\times \Lambda \rightarrow \mathcal{M}$ is to estimate a model $M_\lambda\in \mathcal{M}$, from the space of all models $\mathcal{M}$ with hyperparameters $\lambda\in\Lambda$, that optimizes an objective function, for example  a loss function $\mathcal{L}$,  over a data set distribution $D\in\mathcal{D}$, where $\mathcal{D}$ is the set of all data sets, such that:
\begin{equation}
\mathcal{A}(D,\lambda)=\text{arg min}_{M_\lambda\in\mathcal{M}}\mathcal{L}(M_\lambda,D)\text{.}
\end{equation}
We define $\Lambda=\Lambda_1\times \dots \times \Lambda_P$ as the $P$-dimensional hyperparameter space that can include continuous or discrete values. The resulting model $M_\lambda$ is optimized with respect to its own parameters, if it is parametric, through a series of updates given a specific hyperparameter configuration $\lambda$. However, the trained model $M_\lambda$ might suffer from a generalization error if $\lambda$  is not carefully optimized.
\\
The task of hyperparameter optimization is to identify an optimal hyperparameter configuration $\lambda^* \in \Lambda$ that results in a model $M_{\lambda^*}$ , such that the generalization error on the validation set is minimized:
\begin{equation}
\lambda^* = \text{arg min}_{\lambda\in \Lambda}\mathcal{L}(\mathcal{A}(D^\text{train},\lambda),D^\text{valid})\text{,}
\end{equation}
and $D^\text{train}\bigcap D^\text{valid}=\emptyset$. Treating $M_\lambda$ as a black box does not offer insight into the importance of hyperparameters, and makes tuning the model exhaustive and time-consuming as it must be trained for every configuration before arriving at the optimal one.\\
%In order to avoid unnecessary evaluations of certain hyperparameters, sequential model-based optimization methods, Algorithm \ref{alg:alg2},  model the response surface of an objective function by fitting a probabilistic model $p(\mathcal{L}|D)$, e.g. by a Gaussian process, to previously observed hyperparameters and their corresponding response. An acquisition function, $\alpha : \Lambda \rightarrow \mathbb{R}$, is introduced as a policy that selects new configurations which maximize a selection criterion denoted as the expected improvement. The most commonly used acquisition function is the expected improvement \cite{jones2001taxonomy} (EI) over the best observed response from previously evaluated hyperparameters $\alpha_\text{min}=\text{min}\{\mathcal{L}(M_{\lambda_0},D),\dots,\mathcal{L}(M_{\lambda_n},D)\}$, for $n$ previous hyperparameter trials, as:
%\begin{equation}
%a(\lambda) = \int \text{max}(0,\alpha_\text{min} - \mathcal{L}(M_\lambda,D))\text{d}p(\mathcal{L}|D)
%\end{equation}
By utilizing the uncertainty that comes with Gaussian modeling, the acquisition function maintains a tradeoff between exploitation and exploration, i.e between selecting hyperparameters in an already well-explored are with a high confidence, which might represent a local optimum, or from a previously unexplored area with low confidence but a potentially lower validation loss.
%\begin{algorithm}
%\caption{Sequential Model-Based Optimization}\label{alg:alg2}
%\begin{algorithmic}[1]  
%	\State \textbf{Input}: $\Lambda$ 	 - hyperparameter space, $H$ history of observations, $a$ - acquisition function, $\hat{f}$ - surrogate model, $I$ number of iterations, $M$ - model, $D$ - target data set
%	\State \textbf{Output}: $\lambda^*$ - best hyperparameter configuration for $D$
%		\For{$i\in \{0,1,\dots,I\}$}
%			\State Fit $\hat{f}$ to $H$
%			\State $\lambda_i = \text{arg max}_\lambda a(\lambda,\hat{f}(\lambda,D))$ 
%%			\State Evaluate $\mathcal{L}(M_{\lambda_i}, D)$
%			\If{$f(\lambda_i,D) < f(\lambda^*, D)$}
%			\State $\lambda^* \leftarrow \lambda_i$
%			\EndIf
%			\State $H \leftarrow H \cup (\lambda_i,f(\lambda_i,D))$
%	\EndFor
%\end{algorithmic}
%\end{algorithm}
\\
The sequential decision nature of hyperparameter tuning and the defined objective of maximizing a reward allows us to cast the problem in a reinforcement learning framework, where we replace the acquisition function by Hyp-RL, the proposed reinforcement learning policy.\\

\section{Hyp-RL: Hyperparameter Tuning in a RL Framework}
In this section, we formulate the sequential decision-making task of hyperparameter tuning as a Markov Decision Process (MDP) and describe the model architecure used.
\subsection{MDP Formulation}
A standard reinforcement learning setting is based on an MDP represented by a tuple $\langle S,A,R,\tau\rangle$, with $S$ as the set of all states, $A$ as the action space, a reward function $R: S\times A \rightarrow \mathbb{R}$ and the transition function $\tau : S\times A\times \mathbb{R} \rightarrow S$ that generates a new state in a possibly stochastic or deterministic environment $\mathcal{E}$. The agent interacts with the environment $\mathcal{E}$, by executing an action from $A=\{1,\dots,|A|\}$ with the task of maximizing the expected discounted reward.  The agent's behaviour is governed by a stochastic policy, $\pi: S \rightarrow A$, which computes the true state-action value, as:
\begin{equation}
Q_{\pi}(s,a) = E_{\pi}\left[\sum\limits_{t=0}^{\infty}	 \gamma^tr_t  \;\;|\;\; S_0 = s, A_0 = a\right], 
\end{equation}
where $\gamma \in [0,1]$ represents the discount factor balancing between immediate and future rewards.\\ To solve this sequential decision problem, the optimal policy selects the action that maximizes the discounted cumulative reward, $\pi_{\ast}(s) \in \text{arg max}_a Q_{\ast}(s,a)$ , where $Q_{\ast}(s,a)$ denotes the optimal action value.
One of the most prominent value-based methods for solving reinforcement learning problems is Q-learning \cite{watkins1992q}, which directly estimates the optimal value function and obeys the fundamental identity, known as the Bellman equation \cite{bellman1957functional}:
\begin{equation}
Q_{\ast}(s,a) = E_\pi\left[r + \gamma \max_{a'}Q_{\ast}(s',a')\;\;|\;\; S_0 = s, A_0 = a\right]
\end{equation}
where $s' = \tau(s,a)$.
Hyperparameter optimization can also be formulated as a MDP. As mentioned previously $\Lambda$ is the space of all hyperparameter configurations and the hyperparameter response surface can be defined by the function $f$ as:
\begin{equation}
f : \mathcal{D}\times\Lambda \rightarrow \mathbb{R}
\end{equation}
The response can be any meaningful performance metric, so without loss of generality, we consider it to be the validation loss of the model $M$ trained with hyperparameters $\lambda$:
\begin{equation}
f(D,\lambda) = \mathcal{L}\left(M_\lambda(D^\text{train}),D^\text{valid}\right)
\end{equation}
The agent navigates the hyperparameter response space through a series of actions, which are simply the next hyperparameter configurations to be evaluated, and thus the total number of actions correponds to the total number of possible hyperparameter configurations, i.e. :
\begin{equation}
A = \Lambda
\end{equation}
The state of the environment is defined as the data set metafeatures $\mathbb{D} = \mathbb{R}^{w}$, with $w = \textbf{dim}(\mathbb{D})$ as the number of metafeatures, described in Section \ref{nnMeta}, plus the history of evaluated hyperparameter configurations and their corresponding response: 
\begin{equation}
S=\mathbb{D}\times(\Lambda\times R)^\ast
\end{equation}
The reward function is set as the hyperparameter response function, and depends on the data set $D$ and the action selected, as shown below:
\begin{equation}\label{eq:reward}
R(D,a) = - f(D,\lambda=a)
\end{equation}
considering that the agent's task is to maximize the reward. The observed reward depends solely on the data set and the hyperparameter configuration selected.  Once an action is selected, a new hyperparameter configuration is evaluated. The transition function then generates a new state, $s'$, by appendinding the newly evaluated hyperparameter configuration, $\lambda$, and the corresponding
reward $r$ observed to the previous state $s$:
\begin{equation}\label{eq:transition}
%%   \tau(s,a)= & (d,((\lambda_1,r_1),\dots,(\lambda_t,r_t), (\lambda=a, R(D,\lambda=a)))),
%%   \\ & \text{for } s = (d, ((\lambda_1, r_1), \ldots, (\lambda_t,r_t)))
%% ISSUE: D is not defined.          
  s' = \tau(s,a,r)= (s, (\lambda=a, r)))
\end{equation}
for $s = (d, (\lambda_0, r_0), \ldots, (\lambda_t,r_t))$ and $d = \text{metafeatures}(D) \in \mathbb{D}$.
Inherently, each state $s$ encompasses all previously seen states. The agent reaches a terminal state in two cases, either the agent exceeds a pre-allocated budget $T$, for example running time, or the same action is selected twice in a row. We impose the second condition to ensure that the agent keeps on exploring the hyperparameter space and does not settle on a specific rewarding configuration. The proposed algorithm is summarized in Algorithm \ref{alg:alg1}.
We alternate between data sets, $D$ by randomly sampling the initial state $s_0$ as
$s_0 = (\text{metafeatures}(D), (\{0\}^{\textbf{dim}(\Lambda)},0)), \quad D\sim\text{Unif}(\mathcal{D})$.
% To transition across data sets, we define a fixed number of episodes per data set throughout the training phase after which a new data set is selected.
\begin{algorithm}[t!]
\caption{Hyp-RL}\label{alg:alg1}
\begin{algorithmic}[1]  
  \State \textbf{Input}:
     $\mathcal{D}$ - data sets,
     $\Lambda$ - hyperparameter grid,
     $\gamma$ - discount factor,
     $N_u$ - target update frequency,
     $N_b$ - replay buffer size,
     $N_e$ - number of episodes per data set,
     $T$ - number of actions per episode
		
  \State initialize $\hat Q$ network parameters $\theta$ randomly; $\theta^-= \theta$; replay buffer $\mathcal{B}=\emptyset$
  \For{$N_e \cdot |\mathcal{D}|$ iterations}
    \State $s_0 = (\text{metafeatures}(D), (\{0\}^{\textbf{dim}(\Lambda)},0)), \quad D\sim\text{Unif}(\mathcal{D})$
    \For{$t\in 0,\dots,T$ and while $s_t$ is not terminal}
    % \While{$e\in 0,1,\dots,N_e$}
    % \State Initialize state $s_0 \leftarrow (d,0^{|\Lambda+1|})$
    \State Determine next action $a_t$:
    \begin{equation*}
      a_t = 
      \begin{cases}
        \sim\text{Unif}(\Lambda),       & \text{if}\ p\sim\text{Unif}\left([0,1]\right) < \epsilon \\
        \max_a \hat Q(s_t,a;\theta),    & \text{otherwise}
      \end{cases}			
    \end{equation*}
    \State Receive reward $r_t = R(D,\lambda=a_t)$
             \Comment{eq.~\ref{eq:reward}}
    \State Generate new state $s_{t+1} = \tau(s_t,\lambda_t, r_t)$
             \Comment{eq.~\ref{eq:transition}}
    \State Store $\mathcal{B}\leftarrow\mathcal{B}\cup\left\{\left(s_t,s_{t+1},a_t,r_t\right)\right\}$ and replace oldest tuple if $|\mathcal{B}| > N_b$
    \State Sample and relabel a minibatch $B$ of experiences from the replay buffer:
      \begin{align*}
        B= & \{ (s, a, Q(s,s',a,r)) \mid   (s,s',a,r) \sim \text{Unif}(\mathcal{B}) \}
        \\ \text{with } &
            Q(s,s',a,r) = 
        \begin{cases}
          r,                                                         & \text{if $s'$ is terminal}\\
          r + \gamma \max_{a'} \hat Q_\text{target}(s',a';\theta^-),  & \text{otherwise}
        \end{cases}			
      \end{align*}
    \State Update the $\hat Q$ network by minimizing
               $$\theta = \text{arg min}_{\theta'} \sum_{(s,a,Q)\in B} \left(Q - \hat Q(s,a;\theta')\right)^2$$
    \State Replace target parameters $\theta^- \leftarrow \theta$ every $N_u$ steps
    % \EndIF
    % \EndWhile
    \EndFor
  \EndFor
\State \textbf{return} $\theta$
\end{algorithmic}
\end{algorithm}
\\Wrapping this formulation in a reinforcement learning framework, the
agent starts at a random location in the hyperparameter space of a
random data set and is required to navigate the response surface of a
given model type. The agent explores the environment by selecting the
next best hyperparameter configuration, and receives a reward with
every step until a terminal state reached, i.e until the budget is
exceeded or the same hyperparameter configuration is selected twice in
a row. When the agent runs out of episodes, it is relocated to the
response surface of another data set. For the sake of comparison to RL
in games, we can think of the hyperparameter tuning as a maze-like
environment where every data set represents a different level in the
game.

\subsection{Model Architecture}
\begin{figure}[t]
    \centering
	\subfigure{\includegraphics[width=12cm, height=5cm]{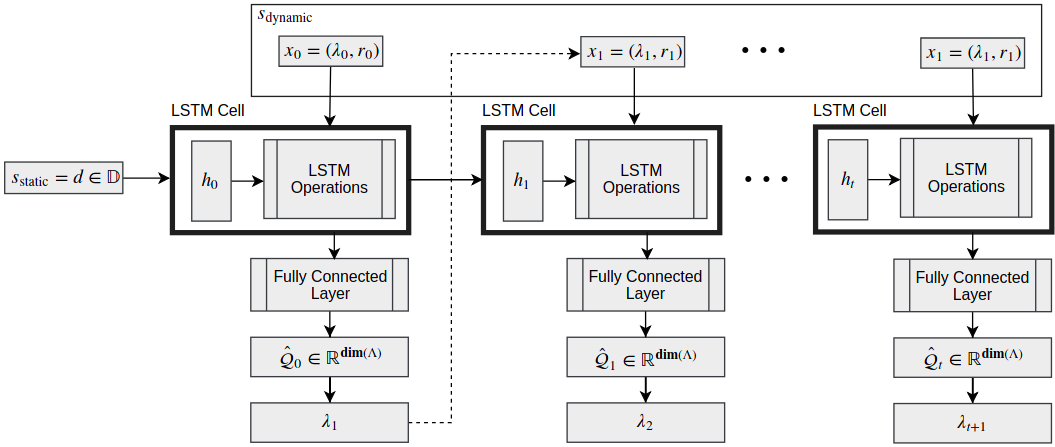}}
	\caption{The schematic illustration of Hyp-RL}
	\label{farch}
\end{figure}

The state representation is decomposed into two parts: the static data set features $s_\text{static} = d$, and the sequence of selected hyperparameter configurations and their correponding rewards, $s_\text{dynamic} \in \left(\Lambda\times R\right)^\ast$, which we model as a dynamic multivariate time series distribution at time $T$ as $s_\text{dynamic} \in \left(\Lambda\times R\right)^T$. Each channel hence in $s_\text{dynamic}$ represents a hyperparameter value, with the final channel including the reward. 
For general purpose sequence modeling, long short term memory (LSTM) cells \cite{hochreiter1997long} have proven to be powerful structures to model long-range dependencies. A simple LSTM cell includes a memory cell, $c_t$ that accumulates information across time steps. The amount of information carried over by $c_t$  is regulated by means of an input gate,  $i_t$, and a forget gate, $f_t$, whereas an output gate $o_t$ controls the final state $h_t$. The key equations that are executed by an LSTM are presented in Equation \ref{lstm}:
\begin{equation}
\begin{aligned}
f_t &= \sigma_g(W_f\cdot[h_{t-1},x_t] +b_f)\\
i_t &= \sigma_g(W_i\cdot[h_{t-1},x_t]+b_i)\\
o_t &= \sigma_g(W_o\cdot[h_{t-1},x_t]+b_o)\\
c_t &= f_t\odot c_{t-1} + i_t\odot(\tanh(W_c[h_{t-1},x_t] + b_c))\\
h_t &= o_t\odot\tanh(c_t)
\end{aligned}
\label{lstm}
\end{equation}
with $b_.$ and $\odot$ denote the bias and the Hadamard product respectively, and $x_t = (\lambda_t,r_t) \in s_\text{dynamic}$. The hyperparameter response surface differs across data sets, and therefore it is important that the policy is conditioned on the data set itself. We take that into consideration by modifying the initialization of the LSTM, where the initial state, $h_0$, is commonly initialized as a the zero vector, i.e. $h_0\in\{0\}^{N_h}$, with $N_h$ as the number of hidden units in the cell. More specifically, we set $h_0$ as: 
\begin{equation}
h_0 = W_0\cdot s_\text{static}
\end{equation}
where $W_0\in\mathbb{R}^{N_h\times \textbf{dim}(\mathbb{D})}$. Through this formulation, the agent is able to start navigating the hyperparameter response surface intelligently from the very start, as opposed to Bayesian-based approaches that require the evaluation of several randomly selected configurations to initally fit the surrogate model. A schematic illustration of our model is presented in Figure \ref{farch}.

\section{Meta Data Set Description: nnMeta}\label{nnMeta}
Meta features are descriptive statistical information about a data set distribution, and are important factors in learning algorithms that are trained across data sets. These indicators can be as simple as a one-hot encoded vector, however this does not entirely capture the latent relationships between different data sets. For this reason, we look into the observations within each data set and compute different statistics, such as skewness, kurtosis, dimensionality, etc., as a richer encoding of every distribution. These features are further scaled to a Gaussian distribution to reduce variations across data sets. The complete list of features are presented in Table \ref{meta}, and have been initially presented in \cite{wistuba2016hyperparameter}.
\begin{table}[htp]
\caption{The list of all metafeatures used to generate nnMeta}    
\centering
\begin{center}
\begin{tabular}{p{2.35in}|l|l}
\toprule
\midrule
Number of Instances& Kurtosis Min\\
Log Number of Instances& Kurtosis Max\\
Number of Features& Kurtosis Mean\\
Log Number of Features& Kurtosis Standard Deviation\\
Data Set Dimensionality& Skewness Min\\
Log Data Set Dimensionality& Skewness Max\\
Inverse Data Set Dimensionality& Skewness Mean\\
Log Inverse Data Set Dimensionality& Skewness Standard Deviation\\
\bottomrule
\end{tabular}
\end{center}
\label{meta}
\end{table}
%\vspace{-0.5cm}%\subsection{Collecting the meta-feature}
To generate the meta-data set, we randomly select 50 UCI classification data sets, and train a neural network architecture with different hyperparameter configurations in Keras \cite{chollet2015keras}. The hyperparameter response is obtained by evaluating on 20\% of the data after training on the remaining 80\%. Three different groups of hyper-paramaters are investigated to generate the meta-data set: structure-based which can be summarized by the number of layers, number of neurons and activation function; optimization-based which include the type of optimizer and the number of epochs; and regularization-based which include dropout rate, regularization technique and the regularization constant. The values of the hyperparameters are summarized in Table \ref{hyper}. The experimental results represent the average of a 5-fold based cross-validation. The hyperparameter grid results in 2916 distinct experiments per data set, and 13 dimensions per configuration.
\begin{table}[t!]
\caption{Hyperparameter grid of nnMeta}

    \begin{tabular}{lllllllll}
        \cline{4-9}
        \multicolumn{1}{c}{} & \multicolumn{2}{l}{} & \multicolumn{3}{|l}{Values} & \multicolumn{3}{|c|}{Encoding} \\
        \cline{2-9}

        \multicolumn{1}{l}{\multirow{8}{*}{}} & \multicolumn{1}{|l}{\multirow{3}{*}{Structure}} & \multicolumn{1}{|l}{Activation Function} & \multicolumn{1}{|l}{ReLU} & \multicolumn{1}{|l}{leakyReLU} & \multicolumn{1}{|l}{tanh} & \multicolumn{3}{|c|}{One-hot} \\ 
        
        \cline{3-9}
        \multicolumn{1}{l}{\multirow{8}{*}{}} & \multicolumn{1}{|l}{\multirow{3}{*}{}} & \multicolumn{1}{|l}{Number of Neurons} & \multicolumn{1}{|l}{5} & \multicolumn{1}{|l}{10} & \multicolumn{1}{|l}{20} & \multicolumn{3}{|c|}{Scalar} \\ 
        \cline{3-9}
        \multicolumn{1}{l}{\multirow{8}{*}{}} & \multicolumn{1}{|l}{\multirow{3}{*}{}} & \multicolumn{1}{|l}{Number of Hidden Units} & \multicolumn{1}{|l}{10} & \multicolumn{1}{|l}{20} & \multicolumn{1}{|l}{50} & \multicolumn{3}{|c|}{Scalar} \\ \cline{2-9}
        
        \multicolumn{1}{l}{\multirow{8}{*}{}} & \multicolumn{1}{|l}{\multirow{3}{*}{Optimization}} & \multicolumn{1}{|l}{Optimizer} & \multicolumn{1}{|l}{Adam} & \multicolumn{1}{|l}{AdaDelta} & \multicolumn{1}{|l}{AdaGrad} & \multicolumn{3}{|c|}{One-hot} \\ 
        
        \cline{3-9}
        \multicolumn{1}{l}{\multirow{8}{*}{}} & \multicolumn{1}{|l}{\multirow{3}{*}{}}  & \multicolumn{1}{|l}{Number of Epochs} & \multicolumn{1}{|l}{10} & \multicolumn{1}{|l}{100} & \multicolumn{1}{|l}{} & \multicolumn{3}{|c|}{Scalar} \\ 
        \cline{2-9} 
        
        \multicolumn{1}{l}{\multirow{8}{*}{}} & \multicolumn{1}{|l}{\multirow{3}{*}{Regularization}} & \multicolumn{1}{|l}{Dropout} & \multicolumn{1}{|l}{0} & \multicolumn{1}{|l}{0.2} & \multicolumn{1}{|l}{0.4} & \multicolumn{3}{|c|}{Scalar} \\ 
        
        \cline{3-9}
        \multicolumn{1}{l}{\multirow{8}{*}{}} & \multicolumn{1}{|l}{\multirow{3}{*}{}} & \multicolumn{1}{|l}{$L_p$ Regularization} & \multicolumn{1}{|l}{$L_1$} & \multicolumn{1}{|l}{$L_2$} & \multicolumn{1}{|l}{} & \multicolumn{3}{|c|}{One-hot} \\ 
        \cline{3-9}
        \multicolumn{1}{l}{\multirow{8}{*}{}} & \multicolumn{1}{|l}{\multirow{3}{*}{}} & \multicolumn{1}{|l}{Regularization Constant} & \multicolumn{1}{|l}{0.01} & \multicolumn{1}{|l}{0.001} & \multicolumn{1}{|l}{0.0001} & \multicolumn{3}{|c|}{Scalar} \\ \cline{2-9}             
    \end{tabular}
\label{hyper}
\end{table}
\section{Experiments}
The netowork is implemented in Tensorflow \cite{abadi2016tensorflow} and the reported results are the averaged over 5 environments representing 5-splits. The environments are wrapped in the OpenAI framework \cite{brockman2016openai}. The code is available here \footnote{https://github.com/hadijomaa/HypRL.git}.
\subsection{Baselines}
We compare our proposed approach with several adopted baselines that vary in terms of complexity and scalability. 
\begin{enumerate}
\item Random Search \cite{bergstra2012random} (RS): One of the simplest hyperparameter tuning methods, in which several configurations are evaluated at random from the hyperparameter search space, which have shown to outperform manual and grid search.
\item Independent Gaussian Process \cite{snoek2012practical} (I-GP): In this approach, the surrogate is modeled by a Gaussian process with a squared-exponential kernel and automatic relevance determination. Hyperparameter tuning is done on each data set independently.
\item Spearmint \cite{snoek2012practical}: Similar to I-GP, the surrogate is modeled by a Gaussian process with a Mat\'ern 5/2 kernel.
\item Factorized Multi-Layer Perceptron \cite{schilling2015hyperparameter} (F-MLP): Instead of a Gaussian process, this approach models the surrogate model of all data sets by a neural network. To accomplish this task, binary data set indicators are appended to the hyperparameter configuration as the input vector. 
In the input layer, feature interactions are explicitly modeled as the prediction of a factorization machine instead of a linear model. %\item Two-Stage Transfer Surrogate\cite{wistuba2016two} (TST-R): As the name suggest, this approach models the surrogate model in two stages and leverages similarity across data sets. Particularly, in the first stage, a Gaussian process is trained as a surrogate model to estimate the hyperparameter response of a new data set as well as previous data sets. In the second stage the surrogate models are averaged using normalized weights  proportional to the similarity of the new unseen data set features and existing meta-data sets.
\item Hyp-RL: This is the approch proposed by this paper. We train a policy that helps an agent navigate the hyperparameter response surface of several data sets in a way that maximizes a future reward given previous hyperparameter configurations and observed rewards. \\

\end{enumerate}
RS, I-GP, and Spearmint cannot carry over information from previous experiments, i.e. on other data sets, whereas F-MLP is a strong baseline that leverages meta-knowledge to reconstruct a scalable hyperparameter response surface. I-GP, Spearmint and F-MLP follow the sequential model-based optimization approach.% detailed in Algorithm \ref{alg:alg2}. 
\\Our hyperparameter finding controller has its own set of hyperparameters, too, so we performed a limited grid search over the following parameters: the number of hidden units in the LSTM cell $N_\text{cells}\in\{16,32\}$, the number of neurons for the single fully connected layer $N_\text{layer}\in\{32,64\}$, replay buffer size $N_b\in\{10000,50000\}$, target update frequency $N_u\in \{500,1000\}$, training frequency $N_\text{train}\in\{4,8\}$ and finally learning rate $lr \in \{0.001,0.0001\}$ experiments. 

\subsection{Evaluation Metrics}
We follow the evaluation metrics of the state-of-the-art papers
\cite{wistuba2016two}. For the average rank (AR), we rank the
performance of the best hyperparameter configuration obtained by a
tuning strategy with respect to other methods and then average over
all data sets. This highlights the difference between different
approaches, however it does not offer insight into how well the
selected hyperparameters are with respect to the global optimum. For
that, we use the average distance to the minimum (ADTM) as the second
evaluation metric. After $t$ trials, ADTM is defined as
\begin{align*}
 \text{ADTM}((\Lambda_t^D)_{D\in\mathcal{D}},\mathcal{D}) = \frac{1}{|\mathcal{D}|}\sum\limits_{D\in\mathcal{D}}
     \min_{\lambda\in\Lambda_t^D}\frac{f(D,\lambda)-f(D)^\text{min}}{f(D)^\text{max}-f(D)^\text{min}}
\end{align*}
with $\Lambda_t^D$ as the set of hyperparameters that have been selected by a
hyperparameter optimization method for data set $D$ in the first $t$ trials
and $f(D)^{\min},f(D)^{\max}$ the range of the loss function on the hyperparameter grid
$\Lambda$ under investigation.
%\subsection{Evaluation Metrics}
%We follow the evaluation metrics of the state-of-the-art papers \cite{wistuba2016two}. For the average rank (AR), we rank the performance of the best hyperparameter configuration obtained by a tuning strategy with respect to other methods and then average over all data sets. This highlights the difference between different approaches, however it does not offer insight into how well the selected hyperparameters are with respect to the global optimum. For that, we use the average distance to the minimum  (ADTM) as the second evaluation metric. After $t$ trials, ADTM is defined as 
%\[ ADMT(\Lambda_t,\mathcal{D}) = \frac{1}{|D|}\sum\limits_{D\in\mathcal{D}}\min_{\lambda\in\Lambda_t}\frac{f(D,\lambda)-f(D)^\text{min}}{f(D)^\text{max}-f(D)^\text{min}}\]
%with $\Lambda_t$ as the set of hyperparameters that have been evaluated in the first $t$ trials.
\subsection{Results and Discussion}
Before discussing the performance of the proposed approach for the task hyperparameter tuning, we investigate the learning progress of the RL policy. 
\begin{figure}[t!]
   \centering
\subfigure[Average Q Values across Splits]{\label{fig:as1}\includegraphics[width=6cm, height=6cm]{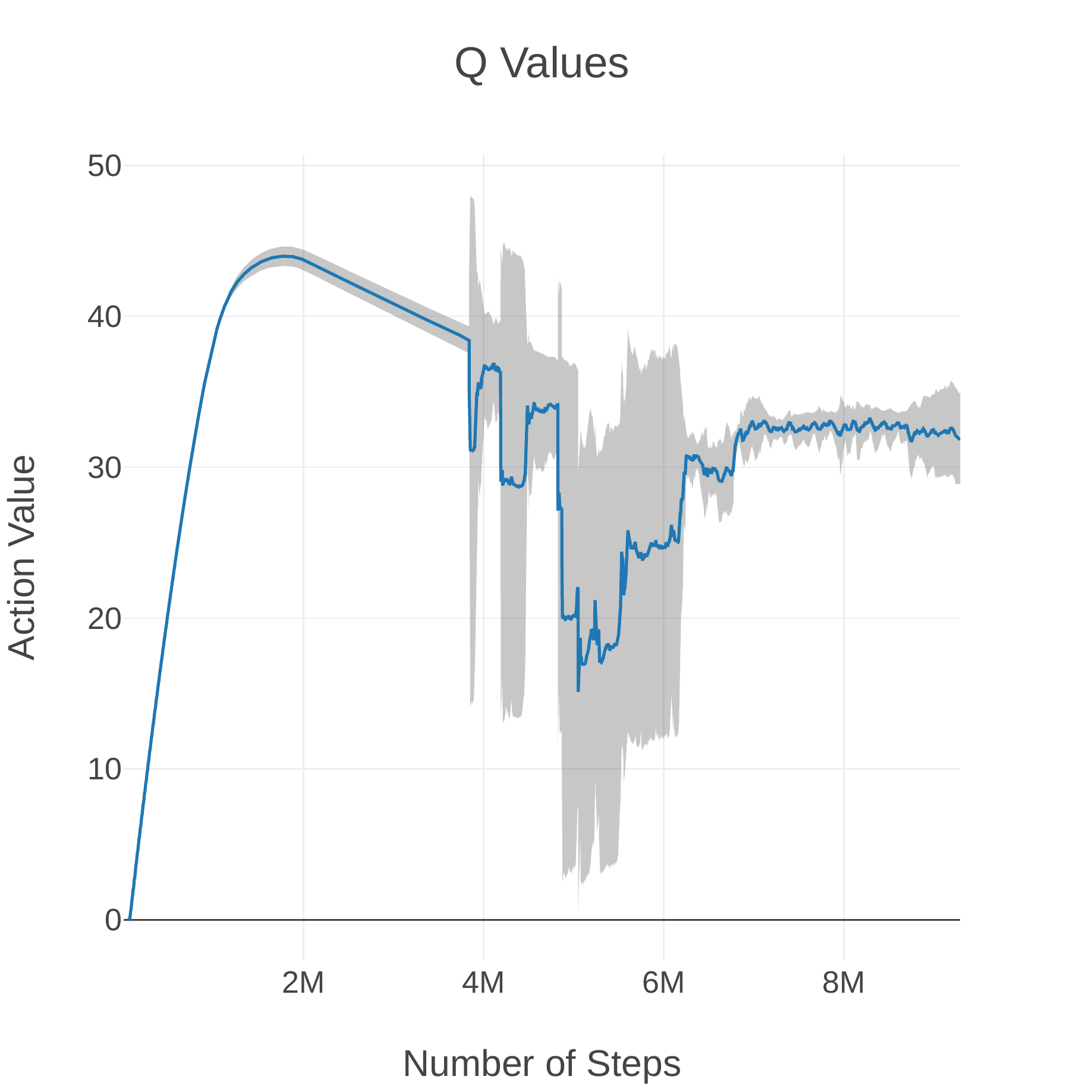}}
\subfigure[Average Number of Trials across Splits]{\label{fig:as2}\includegraphics[width=6cm, height=6cm]{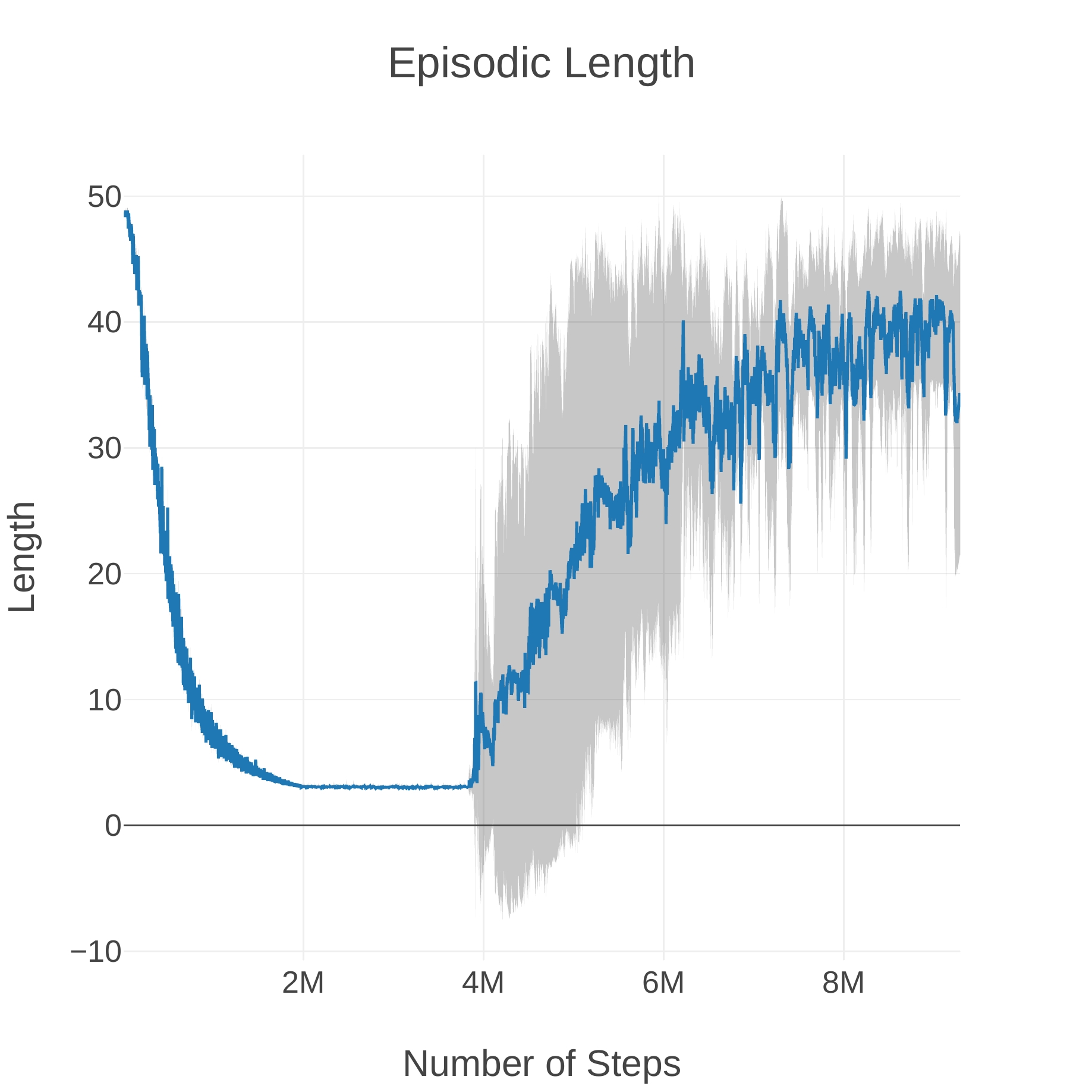}}
\caption{Learning progress of the proposed policy}
   \label{fig}
\end{figure}
\subsubsection{Evaluating the RL Policy}
\begin{figure}[t!]
    \centering
	\subfigure[Average EI per Episode]{\label{fig2:a}\includegraphics[width=6cm, height=6cm]{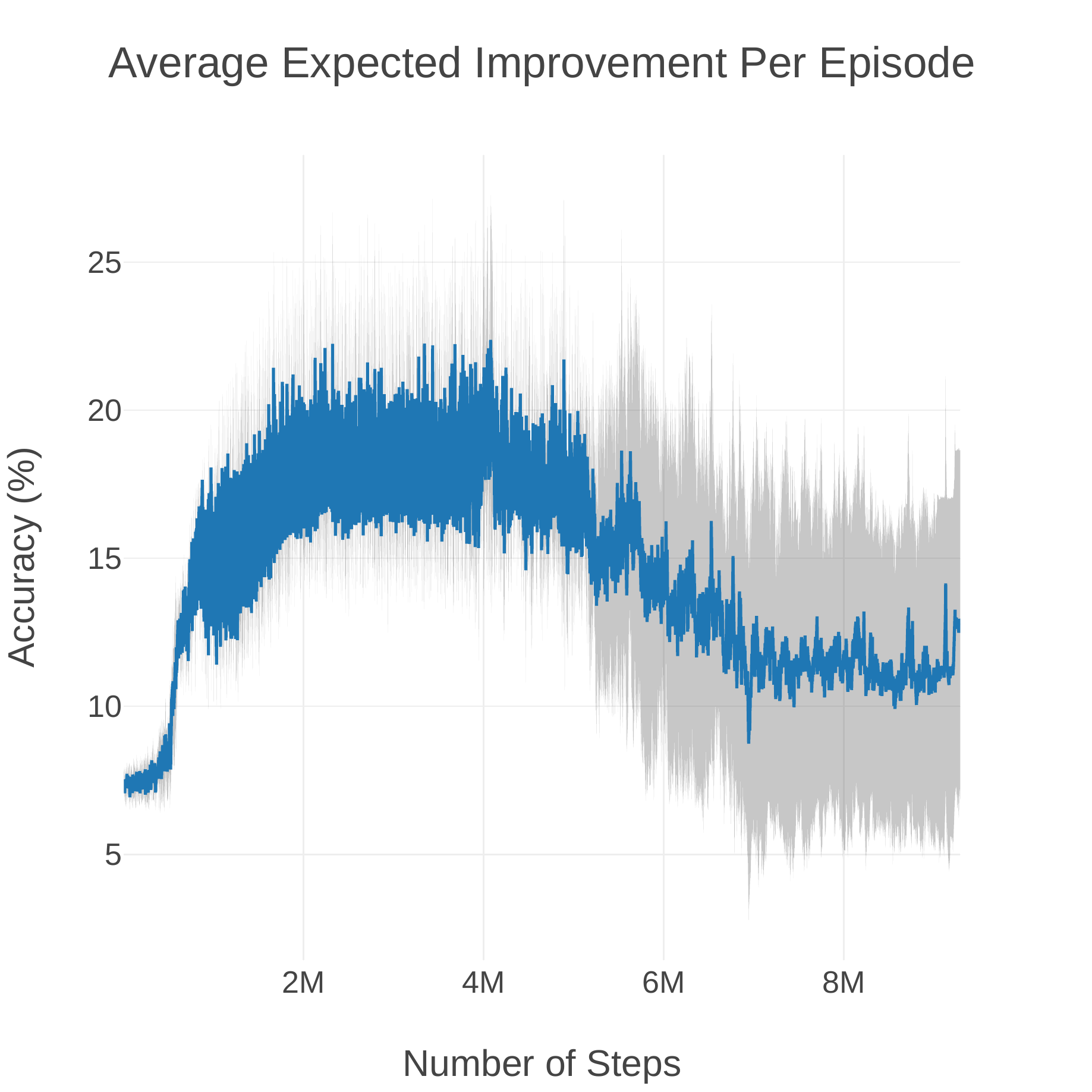}}
	\subfigure[Average Observed Reward per Episode]{\label{fig2:b}\includegraphics[width=6cm, height=6cm]{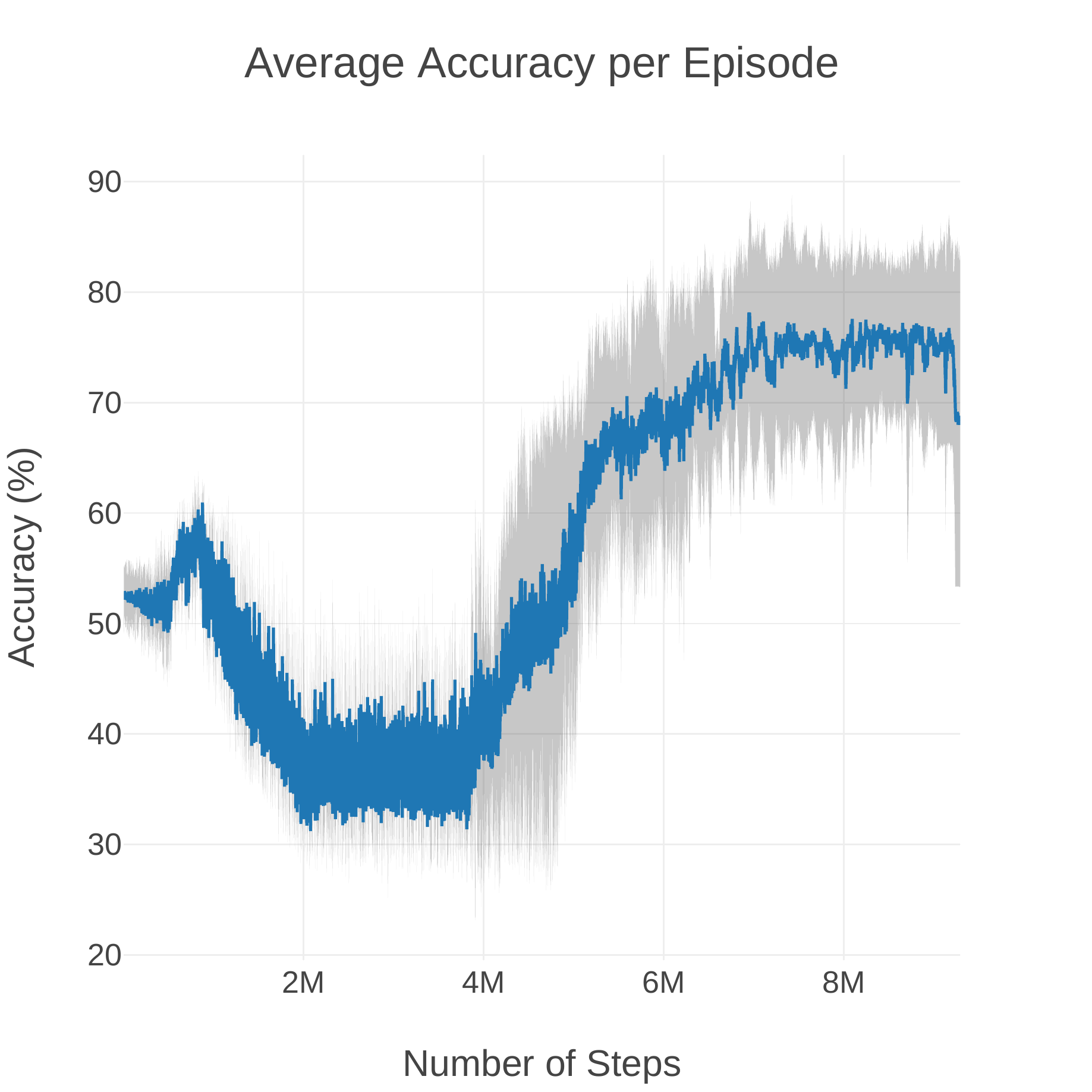}}
	\caption{Performance of the proposed policy}
	\label{fig2}
\end{figure}

The learning curves of our RL policy are presented in Figures \ref{fig} and \ref{fig2}. For the first several hundred episodes, no actual learning takes place as the agent is left free to randomly explore, i.e. exploration ratio $\epsilon$ is relatively high, until the buffer $\mathcal{B}$ is completely filled. We can clearly see from the short episodic length in the early stages, Figure \ref{fig:as2} that the agent tends to terminate quickly due to the selection of similar hyperparameter configurations consecutively, which can be attributed to the random initialization of the model. However as learning progresses, we notice the episode length, and respectively the episodic reward increases, which highlights the balance between exploration, i.e. selecting new (unseen) actions, hyperparameter configurations, and exploitation, i.e. selecting actions that result in a high reward.  We also notice that the action-value increases, which reflects the growing confidence of the model with each action selection. The observed overestimation is a typical issue in function estimation based on Q-learning \cite{thrun1993issues}, and has been addressed by several approaches \cite{hessel2018rainbow,van2016deep} and can be considered future work for this type of formulation.\\
%Obviously the episodic reward has an upper bound limited to the hyperparameter response surface. In an optimal scenario where the selection of the same configuration twice in a row is forbidden, we can calculate the upper bound, represented by the dashed line in Figure \ref{fig1}, as the sum of the rewards obtained from the two best performing configurations, selected alternatively, throughout the trials, and averaged over the training meta-data set over the different folds.The cumulative episodic reward gradually approaches this upper bound as the learning progresses.The temporal difference also decreases and plateaus before the sudden increase in the observed episodic reward.
As the number of trials increases, the aggregated episode reward also increases, however a detailed look at Figure \ref{fig2:b} shows the average reward per trial and respectively reflects that improved navigation of the policy across hyperparameter response surfaces. The expected improvement (EI) is also investigated. As mentioned earlier, Gaussian-based optimization techniques employ EI as an acquisition function that estimates the performance of hyperparameters before selecting the one with the highest potential. Our reinforcement-learning agent inherently models this effect, as initially the EI overshoots, around the time when the observed reward is below the global optimum, and then as the reward increases, EI naturally decreases as there can only be so much improvement to be achieved as the agent arrives closer at the best configuratiion.
The policy network has 115369 parameters and required 24 GPU hours to train for 10 million frames. It is worth mentioning, however that online hyperparameter optimization is immediate as the configurations are selected through via forward pass. \begin{wrapfigure}{r}{5.5cm}
\caption{Average time (in seconds) to finish one trial; The y-axis is plotted in log-scale}\label{wrap-fig:1}
\includegraphics[width=5.5cm]{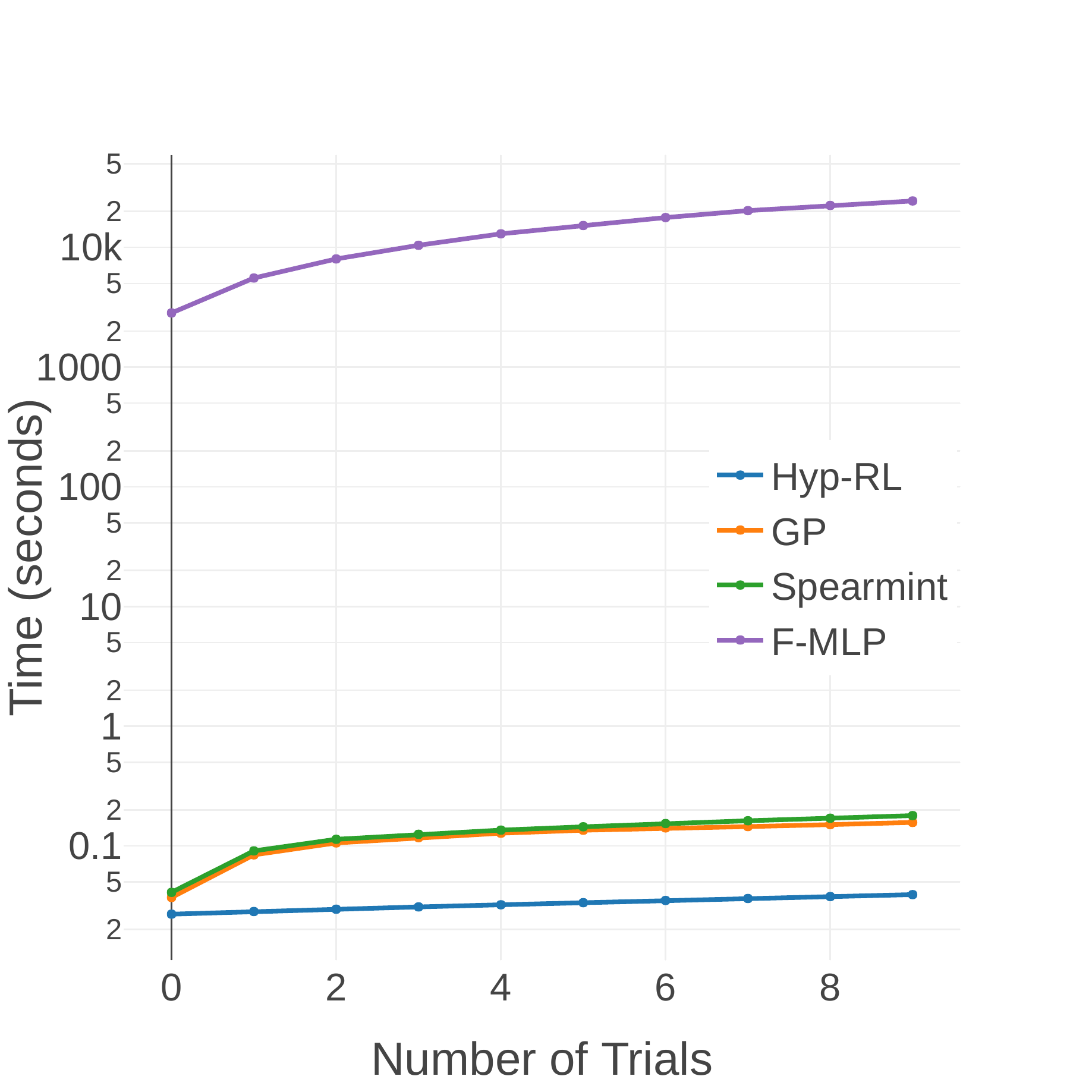}
\end{wrapfigure}

\begin{figure}[t]
    \centering
	\subfigure[Average Distance to the Minimum]{\label{fig3:a}\includegraphics[width=6cm, height=6cm]{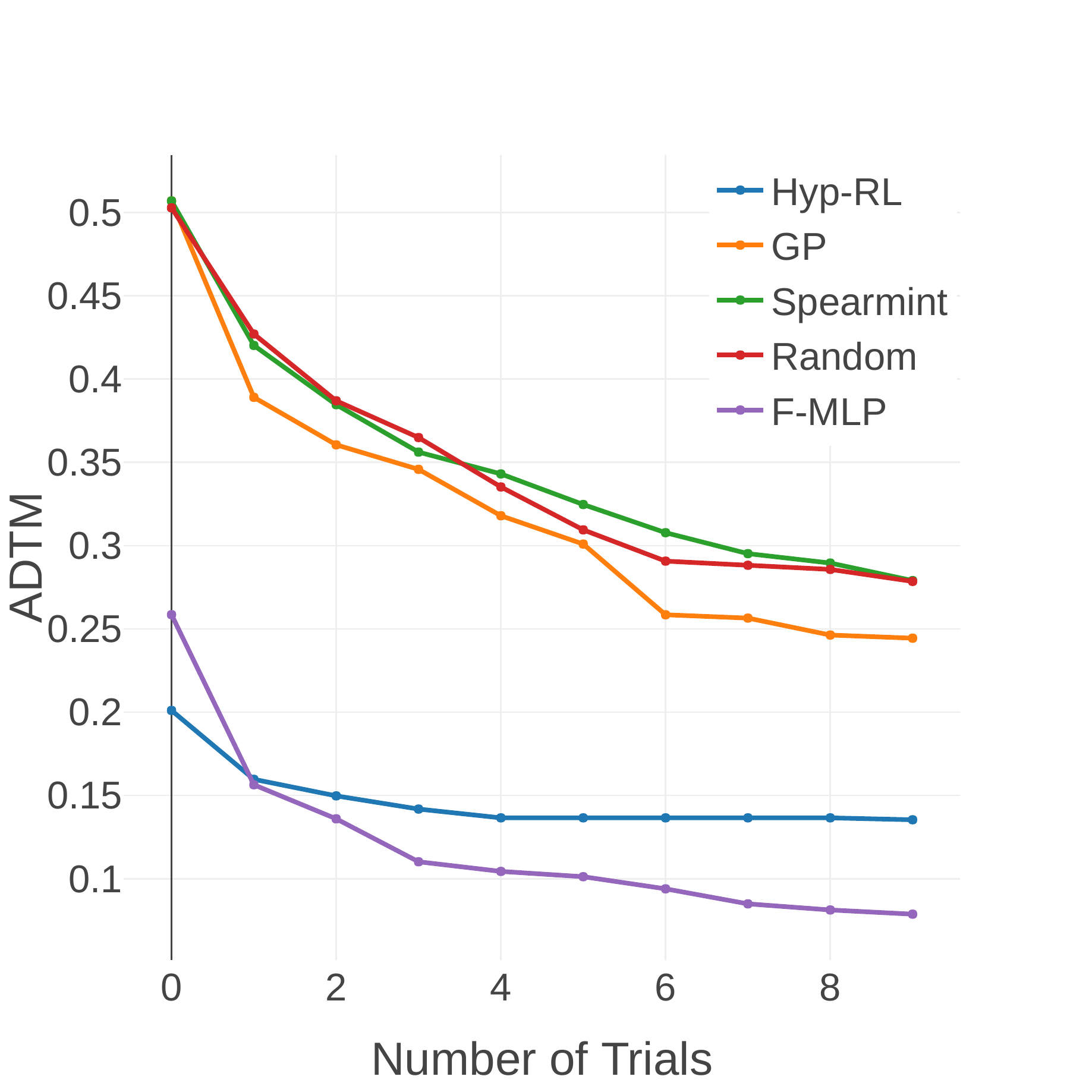}}
	\subfigure[Average Rank]{\label{fig3:b}\includegraphics[width=6cm, height=6cm]{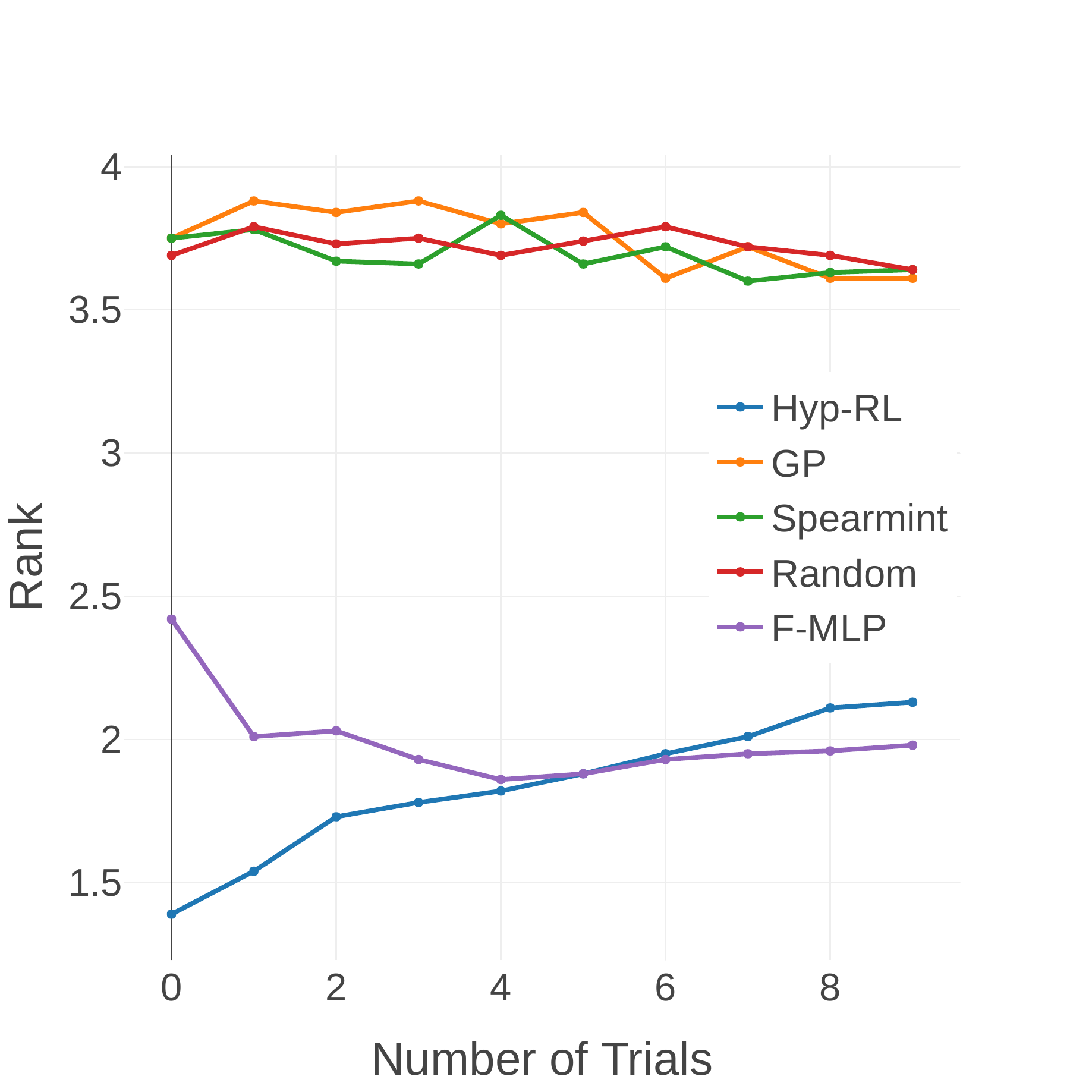}}
%	\subfigure[Average time (in seconds) to finish one trial; The y-axis is plotted in log-scale]{\label{fig3:c}\includegraphics[width=5cm, height=6cm]{time.pdf}}
	\caption{Hyp-RL consistently outperforms the baselines that do not make use of knowledge transfer. Hyp-RL demonstrates competitive performance against F-MLP in a much more efficient approach.}
	\label{fig3}
\end{figure}
\subsubsection{Hyperparameter Tuning} Now that we have established that the RL formulation is capable of learning a meaningful policy to navigate different hyperparameter response surfaces, we can move on to hyperparameter tuning on unseen data sets. We investigate the performance of the different optimization methods for the problem of hyperparameter tuning and summarize the results in Figure \ref{fig3}. The acquisition function of the Gaussian-based optimization techniques are usually initialized with randomly evaluated hyperparameter configurations before they start selecting configurations in a meaningful way. However, our policy, which is conditioned on the data metafeatures does not suffer from this problem. From the very beginning, Hyp-RL selects hyperparameter configurations with better performance than the rest, which is indicative of the capability of the proposed policy to scale across data sets. The proposed method also demonstrates competetive performance with respect to F-MLP, which relies heavily on knowledge transfer.
It is worth mentioning that since our agent does not learn a surrogate model, inference becomes significantly faster, as opposed to SMBO-based techniques which re-fit a surrogate model after every configuration selection. This additional step, along with the curse of dimensionality, which renders Gaussian processes' complexity cubic in the number of data points, are avoided using our formulation. Even F-MLP, which trains a neural network as a surrogate model, demands a large time budget as compared to other approaches, as seen in Figure \ref{wrap-fig:1}.

%\subsubsection{Reproducability}
%The code used to generate the results is hosted by the University of Hildesheim \footnote{http://147.172.223.201/oc/index.php/s/LkNTWwnPrdyQWrz} which the reviewers can access using this password:\\"pyhidah9102". We will push our code to GITHUB if the paper gets accepted. We set up 5 environments representing the 5-splits that we have used and wrapped the environments in the OpenAI framework. %We will also provide the metafeatures of the data set used. %To ensure that our policy is exposed to all data sets uniformly, we iteratively loop over the data sets after a fixed number of episodes. We also use an a linear decaying exploration function which adjusts $\epsilon$ w.r.t number of steps.

\section{Conclusion}
In this paper, we presented a new approach for hyperparameter optimization within a reinforcement learning framework. We describe a novel environmental setup represented by the history of observed hyperparameter configurations and their corresponding rewards, and train an agent to select the next hyperparameter setting to be evaluated in a way that maximizes the total reward within a limited budget. The idea extends naturally from Bayesian-based approaches that use an acquisition function as a policy. The proposed approach does not suffer from the cubic dimensionality problem that face Gaussian-based approaches, however the number of actions is still proportional to the grid size. Empirically, our method demonstrates competitive performance w.r.t. the state-of-the-art, especially with a smaller budget, where we notice that from the very first selection, the average distance to the minimum is small. We also show that the agent balances between exploration and exploitation by enforcing termination when an action in an episode is repeated. \\
%Moving forward, we plan to reduce the dimensionality of the action space considerably, by defining the actions as the hyperparameter values, and not hyperparameter configurations. This should be facilitated by training continuous policies. 
\subsection*{Acknowledgment}
We acknowledge the funding provided by the "Zentrales Innovationsprogramm Mittelstand" of the German Federal Ministry for
Economic Affairs and Energy through the project ADDA.% The authors thank Nicolas Schilling (Uni. Hildesheim) for the valuable advice.
\small
\bibliographystyle{splncs04}
\bibliography{jomaa2019b-ecml}
\end{document}